\documentclass[journal]{IEEEtran}
\usepackage{bm}
\usepackage{subfigure}
\usepackage{epsfig}
\usepackage{latexsym}
\usepackage{amsmath}
\usepackage{amsthm}
\usepackage{amssymb}
\usepackage{amsfonts}
\usepackage{color}
\usepackage{enumerate}
\usepackage{multirow}
\usepackage{booktabs}
\usepackage{epstopdf}
\usepackage{algorithm}
\usepackage{algpseudocode}
\usepackage{threeparttable}
\usepackage{diagbox}
\usepackage{lettrine}
\usepackage{colortbl}
\usepackage{booktabs}
\usepackage{graphicx}
\usepackage{setspace}
\usepackage{makecell}
\usepackage[framemethod=TikZ]{mdframed}
\usepackage{pdfpages}
\ifCLASSINFOpdf
\else
\fi
\begin{document}
\begin{figure*}[t]
\centering
\setlength{\abovecaptionskip}{-5pt}
\begin{center}
\includegraphics[width=7.0in]{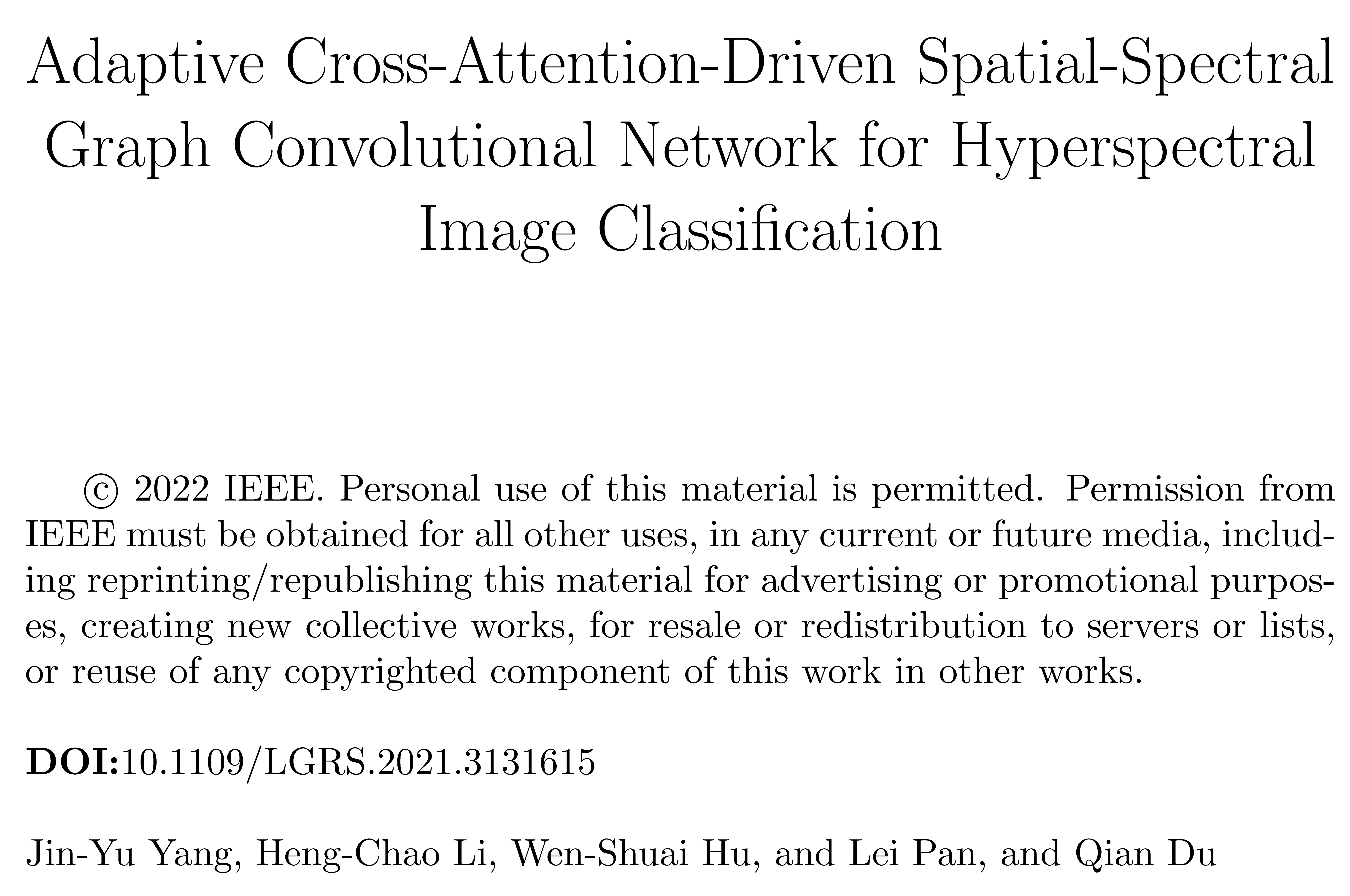}
\end{center}
\tiny
\centering
\end{figure*}
\clearpage
\title{Adaptive Cross-Attention-Driven Spatial-Spectral Graph Convolutional Network for Hyperspectral Image Classification}
\author{
\thanks{This work was supported by the National Natural Science Foundation of China under Grants 61871335 and 62001437. (Corresponding author: Heng-Chao Li.)}
Jin-Yu~Yang,
        Heng-Chao~Li, \emph{Senior Member, IEEE},
        Wen-Shuai~Hu,
        Lei~Pan,
        and Qian~Du, \emph{Fellow, IEEE}
\thanks{Jin-Yu Yang, Heng-Chao Li, and Wen-Shuai Hu are with the School of Information Science and Technology, Southwest Jiaotong University, Chengdu 611756, China (e-mail: lihengchao\_78@163.com).}
\thanks{Lei Pan is with the Southwest Institute of Electronic Technology, Chengdu 610036, China.}
\thanks{Qian Du is with the Department of Electrical and Computer Engineering, Mississippi State University, Mississippi State, MS 39762 USA.}
}
\markboth{IEEE GEOSCIENCE AND REMOTE SENSING LETTERS}%
{Shell \MakeLowercase{\textit{et al.}}: Bare Demo of IEEEtran.cls for IEEE Journals}

\maketitle

\begin{abstract}
 Recently, graph convolutional networks (GCNs) have been developed to explore spatial relationship between pixels, achieving better classification performance of hyperspectral images (HSIs). However, these methods fail to sufficiently leverage the relationship between spectral bands in HSI data. As such, we propose an adaptive cross-attention-driven spatial-spectral graph convolutional network (ACSS-GCN), which is composed of a spatial GCN (Sa-GCN) subnetwork, a spectral GCN (Se-GCN) subnetwork, and a graph cross-attention fusion module (GCAFM). Specifically, Sa-GCN and Se-GCN are proposed to extract the spatial and spectral features by modeling correlations between spatial pixels and
between spectral bands, respectively. Then, by integrating attention mechanism into information aggregation of graph, the GCAFM, including three parts, i.e., spatial graph attention block, spectral graph attention block, and fusion block, is designed to fuse the spatial and spectral features and suppress noise interference in Sa-GCN and Se-GCN. Moreover, the idea of the adaptive graph is introduced to explore an optimal graph through back propagation during the training process. Experiments on two HSI data sets show that the proposed method achieves better performance than other classification methods.
\end{abstract}

\begin{IEEEkeywords}
Hyperspectral image classification, graph convolutional networks, feature extraction, attention mechanism.
\end{IEEEkeywords}

\IEEEpeerreviewmaketitle

\section{Introduction}
\lettrine[lines=2]{H}{yperspectral} images (HSIs) are the 3-D data recording the spatial-spectral information of land covers. HSI classification, as an active research area, has received extensive attentions in many fields. Recently, deep learning methods exhibit good classification performance in HSIs \cite{YC2016,HP2021}.

Graph theory is an effective manner to represent the similar relationship of HSI data for revealing their intrinsic geometric properties.
Specifically, graph embedding methods have been used to learn graph representation in the low-dimensional space for reducing the redundant information of HSI data \cite{FL2016,FL2020}. Inspired by the success of deep learning \cite{AS2022},
Kipf \emph{et al.} \cite{TN2017} proposed a graph convolutional network (GCN) to aggregate and transform the node feature with the neighborhood structure of graph data. Moreover, aiming at the case of limited labeled samples, a spectral-spatial GCN \cite{AQ2019} was utilized to construct a semisupervised framework for HSI classification. Subsequently, Wan \emph{et al.} proposed a multiscale dynamic GCN (MDGCN) \cite{SW2020} and a dual interactive GCN (DIGCN) \cite{SS2021} to capture the spatial information at different scales. A minibatch GCN model was developed to reduce the complexity of training \cite{DF2020}. Recently, the graph topological consistent was utilized to learn the underlying spatial context information of HSIs \cite{YY2021}.

In \cite{JY2020}, GCN was applied to model the temporal correlation between different frames in video. Similar to the temporal relation, the spectral correlation exists in HSI data, which has been applied for HSI classification \cite{HC2020}. Moreover, to produce high-resolution HSI, the spectral structures in low-spatial-resolution HSI and high-spatial-resolution multispectral images were inherited according to the construction of spectral graph along spectral dimension \cite{KM2018}. Therefore, graph learning or GCN might be one of the methods to capture the spectral correlation in the spectral domain.

In recent years, with the development of attention mechanism in deep learning, it has been gradually applied for GCNs, which enables them more learnable with stronger generalization capability. Pu \emph{et al.} \cite{CH2021} designed a multiscale attention mechanism by distinguishing the importances of different spectral bands. Moreover, by allocating weights jointly in the horizontal and vertical directions, the effectiveness of cross attention was demonstrated in \cite{WC2020}.

Nevertheless, there are still some issues in the GCN-based models for HSI classification. Firstly, the initial adjacent matrix may fail to reveal the effective spatial relationship. Secondly, the relationship of spectral bands is underused. Thirdly, existing GCNs usually utilize the addition, multiplication, or concatenation operations to fuse features, which fail to explore the complementary between features. As such, in this letter, an adaptive cross-attention-driven spatial-spectral graph convolutional network (ACSS-GCN) is proposed to jointly extract the spatial-spectral features for HSI classification. The contributions are summarized as follows.
\begin{figure*}[t]
\centering
\setlength{\abovecaptionskip}{-5pt}
\begin{center}
\includegraphics[ width=5.2in]{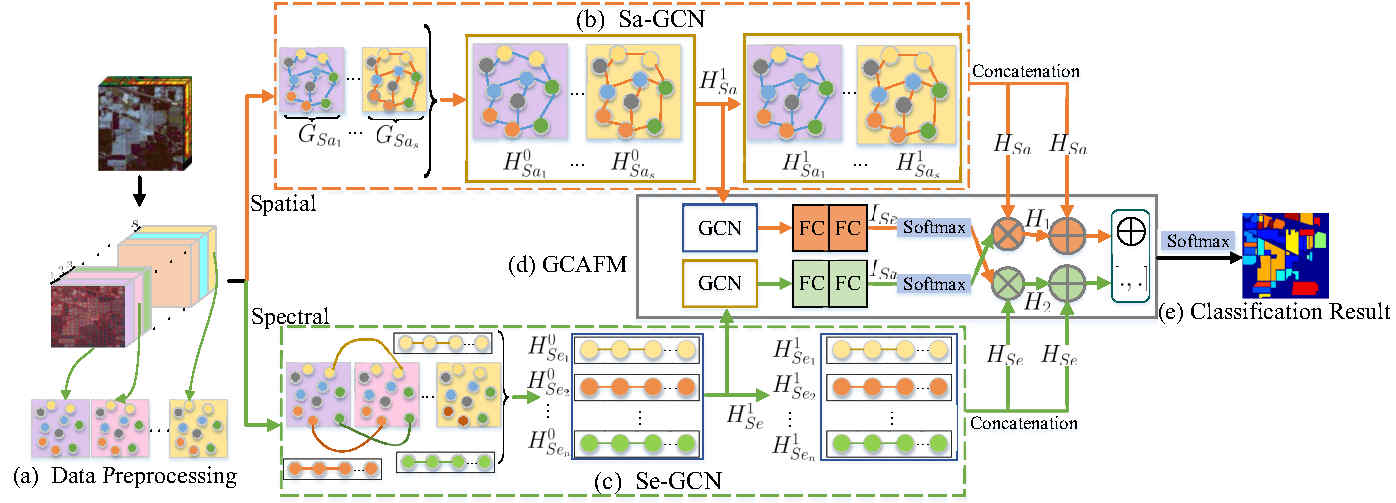}
\end{center}
\tiny
\centering
\caption{The framework of the proposed ACSS-GCN model. (a) Data preprocessing. (b) Spatial graph convolutional network (Sa-GCN). (c) Spectral graph convolutional network (Se-GCN). (d) Graph-based cross-attention fusion module (GCAFM). (e) Classification result. Particularly, ${\oplus}$ and ${\otimes}$ are addition and multiplication operations, respectively. \textcolor[rgb]{0.75,0.56,0.00}{
{{\boxed{ \quad \quad }}}} represents the spatial graph convolution layer. \textcolor[rgb]{0.19,0.33,0.59}{${\boxed{ \quad \quad }}$} is the spectral graph convolution layer. }
\end{figure*}

\hangindent 2.5em
(1) Considering the spatial-spectral characteristic of HSI data, a dual-branch GCN-based spatial-spectral structure is proposed, containing the spatial GCN (Sa-GCN) and spectral GCN (Se-GCN) subnetworks to extract the spatial and spectral features by exploring correlations from the spatial and spectral dimensions, respectively.

\hangindent 2.5em
(2) A novel graph-based cross-attention fusion module (GCAFM) with a spatial graph attention block (SAGB) and a spectral graph attention block (SEGB) is developed by integrating attention mechanism into information aggregation over graph-structured data to explore the complementary information of Sa-GCN and Se-GCN.

\hangindent 2.5em
(3) With the idea of the adaptive graph, a novel ACSS-GCN framework is constructed for the spatial-spectral feature extraction and classification of HSIs.

Experimental results demonstrate the superiority of the proposed method over existing GCNs in HSI classification.

\section{Methodology}

The overall architecture of our ACSS-GCN framework is depicted in Fig. 1. Fig. 1(a) shows data preprocessing on the original HSI data. The backbone of the proposed framework is composed of Sa-GCN and Se-GCN, as shown in Fig. 1(b)-(c), which can capture the spatial and spectral information from HSI data. Moreover, Fig. 1(d) displays the GCAFM, followed by the classification result in Fig. 1(e).
\subsection*{A. Data Preprocessing}
 Simple linear iterative clustering \cite{SW2020} and principal component analysis (PCA) are imposed on the original HSI data to reduce the computational complexity, shown in Fig. 1(a). The feature $x$ of each superpixel is defined by the average of its all pixels, and the features of all superpixels in the HSI data are denoted by $X {=\{x_1,x_2,...,x_n\}\in{\mathbb R}^{n \times s}}$, where $n$ and $s$ are the number of superpixels and spectral bands, respectively. A set of superpixels for the ith spectral band is represented as $Z_{Sa_i}=\left\{x_{1i},x_{2i},\ldots,x_{ni}\right\}$ for $i=1,2,\ldots,s$, while a set of spectral bands for the $j$th superpixel is denoted as $Z_{Se_j}=\left\{x_{j1},x_{j2},\ldots,x_{js}\right\}$ for $j=1,2,\ldots,n$.

\subsection*{B. Spatial-Spectral Feature Extraction}
From Fig. 1, for our ACSS-GCN framework, a dual-branch GCN-based spatial-spectral structure forms its backbone, which is composed of Sa-GCN and Se-GCN to model the correlations between spatial pixels and between spectral bands. Specifically, Sa-GCN extracts spatial features by exploring the spatial relationship, while Se-GCN learns spectral information by modeling the spectral correlation.

\emph{1) Sa-GCN}

The relationship of pixels is distinctive in multiple spectral bands. As shown in Fig. 1(b), a spatial graph corresponding to each spectral band is constructed individually to explore the suitable spatial relationship. For convenience, a set of spatial graphs is written as $\left\{G_{Sa_1},G_{Sa_2},\ldots,G_{Sa_s}\right\}$. The spatial graph $G_{Sa_i}$ of the $i$th spectral band is defined as $(V_{Sa_i},E_{Sa_i},A_{Sa_i})$, where each superpixel is treated  as a graph node, $V_{Sa_i}$ is a set of graph nodes, and $E_{Sa_i}$ is a set of edges.

$A_{Sa_i}$ is the adjacent matrix, indicating whether each pair of superpixels is connected. The Gaussian distance is employed to measure the pairwise similarity between these superpixels. The weight of graph nodes $x_{mi}$ and $x_{hi}$ in $A_{Sa_i}$ is written as
\begin{equation}
{A}_{Sa_{i_{mh}}} =
\begin{cases}
e^{-\gamma{\left \|x_{mi}-x_{hi}\right \|}^2},  & \emph{if $x_{mi} \in {\mathcal N}(x_{hi})$} \\
0, & \emph{otherwise}
\end{cases},
\end{equation}
where $\mathcal N(x_{hi})$ is the set of neighbors of $x_{hi}$, which includes all superpixels directly connected to $x_{hi}$. The value of $\gamma$ is empirically set as 0.5 in the experiments.

After that, the Sa-GCN individually processes each spectral band by the corresponding spatial graph, which has two graph convolutional layers with the ${ReLU}$ function. Thus, for the $i$th band, the spatial information is learned by
\begin{equation}
{H_{Sa_i}^{l+1}} ={ReLU}(L_{Sa_i}{H_{Sa_i}^{l}}{W_{Sa_i}^l}),\\
\end{equation}
where $L_{Sa_i}={\tilde{D}_{Sa_i}^{-\frac{1}{2}}}{\tilde{A}_{Sa_i}}{\tilde{D}_{Sa_i}^{-\frac{1}{2}}}$ with ${\tilde{D}_{Sa_i}}$ being the diagonal degree matrix of $\tilde{A}_{Sa_i}=A_{Sa_i}+I$. $H_{Sa_i}^{l}\in{{\mathbb R}^{N \times \frac{F^l}{s}}}$ denotes the spatial information of the $i$th band in the ${l}$th layer with $H_{Sa_i}^{0}=Z_{Sa_i}$ and $F^l$ is the dimension of feature in $l$ layer. ${W_{Sa_i}^l}$ represents the trainable weight for the $i$th band in the ${l}$th layer, and ${ReLU\left(\cdot\right)}$ is the activation function.

Finally, the spatial features in all bands are concatenated as
\begin{equation}
{H_{Sa}} = [H_{Sa_1},H_{Sa_2},\ldots,H_{Sa_s}].\\
\end{equation}

\emph{2) Se-GCN}

Considering the spectral characteristic of HSIs, a Se-GCN subnetwork is further designed. Similar to \cite{KM2018}, a set of initial spectral graphs $\left\{G_{Se_1},G_{Se_2},\ldots,G_{Se_n}\right\}$ is built to model the spectral relation by using \emph{s}-nearest neighbor strategy in Fig. 1(c). Different from the spatial graph, $G_{Se_j}=(V_{Se_j},E_{Se_j},A_{Se_j})$ is the spectral graph of the $j$th superpixel. The construction of $A_{Se_j}$ is similar to (1), however, the similarity calculation is implemented between each pair of bands in the jth superpixel.

Then, the Se-GCN is built by stacking two convolutional layers with the ${ReLU}$ function, in which each superpixel is separately processed to learn spectral information. The output for the jth superpixel is expressed as follows
\begin{equation}
{H_{Se_j}^{l+1}} ={ReLU}(L_{Se_j}{H_{Se_j}^{l}}{W_{Se}^{l}}),\\
\end{equation}
where $L_{Se_j}={\tilde{D}_{Se_j}^{-\frac{1}{2}}}{\tilde{A}_{Se_j}}{\tilde{D}_{Se_j}^{-\frac{1}{2}}}$, ${H_{Se_j}^l}\in{{\mathbb R}^{1 \times {F^l}}}$ is the spectral information of the jth superpixel in the $l$th layer with ${H_{Se_j}^0=Z_{Se_j}}$, and the dimension in $l$th layer of Se-GCN and Sa-GCN is the same. ${W_{Se}^l}$ is the trainable weight in the lth layer shared by all superpixels, which alleviates the overfitting of the proposed network to a certain extent. Finally, similar to (3), by cascading the outputs of all superpixels, spectral features $H_{Se}\in{{\mathbb R}^{N \times {F^2}}}$ are extracted in Se-GCN.

\subsection*{C. GCAFM}
Based on the above analysis, Sa-GCN and Se-GCN can extract different and complementary features by modeling the spatial and spectral correlations of HSIs, respectively. As such, a novel GCAFM is designed by introducing attention mechanism into graph node to fully exploit this complementary information, whose backbone contains an SAGB, an SEGB, and a fusion block, as illustrated in Fig. 1(d).

\emph{1) SAGB}

From Fig. 1(c), Se-GCN only uses the spectral information such that the spatial correlation of HSI data is not effectively explored. Therefore, an SAGB module with a spatial graph convolutional layer and two fully connected (FC) layers is designed in Fig. 1(d). Therefore, spatial attention weights are generated from the first spectral graph convolutional layer of the Se-GCN to promote the induction of effective spatial feature extraction from Sa-GCN. Specifically, spatial graph convolutional layer is applied to learn spatial information, and two FC layers are utilized to acquire more representational information. The feature from the second FC layer is denoted by ${I_{Sa}\in{{\mathbb R}^{N \times {F^2}}}}$, which is further fed into a softmax function to generate the normalized spatial attention map. The spatial-enhanced features $H_1\in{{\mathbb R}^{N \times {F^2}}}$ can be computed as
\begin{equation}
\begin{aligned}
&{I}_{Sa}=f_{FC}(f_{FC}({ReLU}(L_{Sa}{H_{Se}^1}{W_{Sa}})))\\
&H_1={H}_{Sa}{\odot}softmax({I}_{Sa}),
\end{aligned}
\end{equation}
where $H_{Se}^1\in{{\mathbb R}^{N \times {F^1}}}$ is output of the first layer in Se-GCN. $f_{FC}(.)$ is the FC layer, ${\odot}$ is the element-wise multiplication operation, and $softmax\left(\cdot\right)$ function is used for normalization.

\emph{2) SEGB}

The special structure of Sa-GCN makes it contain the primary spectral relationship. Therefore, an SEGB module is built to generate spectral weight coefficients for adaptively selecting the important spectral features from the Se-GCN. From Fig. 1(d), the SEGB contains a spectral graph convolutional layer, two FC layers, and a softmax function, which, however, calculates the spectral attention map along with the spectral dimension from the output of the first spatial graph convolutional layer in Sa-GCN. In addition, the output of the second FC layer is expressed as $I_{Se}\in{{\mathbb R}^{N \times {F^2}}}$. Therefore, the spectral-enhanced features $H_2\in{{\mathbb R}^{N \times {F^2}}}$ are written as
\begin{equation}
\begin{aligned}
&I_{Se}=f_{FC}(f_{FC}({ReLU}(L_{Se}{H_{Sa}^1}{W_{Se}})))\\
&H_2={H}_{Se}{\odot}softmax({I}_{Se}).
\end{aligned}
\end{equation}
where $H_{Sa}^1 \in{{\mathbb R}^{N \times {F^1}}}$ is the features from the first layer in Sa-GCN.

Finally, in the fusion block, the addition and concatenation operations are applied to fuse the spatial and spectral attention features of Sa-GCN and Se-GCN. Inspired by the residual learning, two fusion methods can be written as
\begin{equation}
\begin{aligned}
&H_{add}=(H_1{\oplus}{H}_{Sa}){\oplus}(H_2{\oplus}{H}_{Se})\\
&H_{con}=\left[H_1{\oplus}{H}_{Sa},H_2{\oplus}{H}_{Se}\right],
\end{aligned}
\end{equation}
where ${\oplus}$ is the element-wise addition. $H_{add}\in{{\mathbb R}^{N \times {F^2}}}$ and $H_{con}\in{{\mathbb R}^{N \times 2{F^2}}}$ mean the extracted spatial-spectral features by these two methods.

\subsection*{D. Adaptive Graph Refinement}
The performance of graph convolution operation mainly depends on the quality of graph structure, which is important to explore the optimal graph for HSI classification. Due to the effects of the noise and complex scenes on HSI data, the spatial and spectral graphs initially constructed in (1) are not accurate, which are not completely suitable for HSI classification. As such, the adaptive matrix $W_p$ is designed to dynamically refine these two graphs, and these corresponding adjacency matrices can be updated as
\begin{equation}
A_o=A_{in}+{\beta}W_pA_{in},
\end{equation}
where $A_{in}=A_{Sa_i}$ or $A_{in}=A_{Se_j}$, and ${\beta}$ is a tuning parameter.
Specifically, each spatial graph with its own $W_p$ makes the model more flexible, while all spectral graphs share the same $W_p$ to avoid the overfitting, all of which are randomly initialized. Similar to \cite{SW2020}, by optimizing the cross-entropy loss function with full-batch gradient descent, all $W_p$ together with other trainable parameters of ACSS-GCN can be updated during training, thus producing the refined spatial and spectral graphs and extracting the more effective spatial-spectral features for HSI classification.
\section{Experimental Results}
In this Section, the performance of the ACSS-GCN is validated on two public HSI data sets, i.e., Indian Pines and University of Pavia. The Indian Pines data set recorded northwestern Indiana with the Airborne Visible/Infrared Imaging Spectrometer (AVIRIS) sensor in 1992, which is made up of 145${\times}$145 pixels with 200 bands. The University of Pavia data were captured in the Pavia University in Italy by the Reflective Optics System Imaging Spectrometer (ROSIS) sensor in 2001, consisting of 610${\times}$340 pixels and 103 spectral bands. In the experiments, 30 pixels of each class are selected as the  training set, while 15 pixels will be extracted for the class where the number of samples dose not reach 30. In addition, overall accuracy (OA), average accuracy (AA), and kappa coefficient ($\kappa$) are used to study the performance of all methods.

To evaluate the performance of our proposed method, seven methods, i.e, SVM \cite{CC2011}, 2-D CNN \cite{YC2016}, GCN \cite{TN2017}, FuNet \cite{DF2020}, AMDPCN \cite{CH2021}, MDGCN \cite{SW2020}, and DIGCN \cite{SS2021}, are utilized for comparison. Moreover, our ACSS-GCN model with two fusion strategies (i.e., addition and concatenation) are represented as ACSS-GCN-A and ACSS-GCN-C, respectively.

\subsection*{A. Parameter Settings}
In the experiments, the spectral dimension after PCA is set to 20. In addition, $dropout=0.5$ for each graph convolutional layer to alleviate the overfitting. For Sa-GCN and Se-GCN, the dimensions of two graph convolutional layers are set as 40 and 20, respectively. The output dimensions of the graph convolutional layer and two FC layers in the SEGB and SAGB modules are set to 40, 25, and 20. Moreover, a FC layer and a softmax function are applied to classify the HSI samples. To eliminate the deviation caused by random initialization, the average value after 5 repetitions is given for each quantitative metric. The epoch is set to 3000. The adaptive momentum optimizer with a learning rate of 0.005 is used to optimize the loss function of cross entropy for the proposed model.

\begin{table}[H]
   \centering
   \setlength{\abovecaptionskip}{-2pt}
   \renewcommand\thetable{\Roman{table}}
   \renewcommand\tabcolsep{3.0pt}
   \caption{Sensitivity Analysis under Different Values of Parameter ${\beta}$}
    \scriptsize
     \begin{tabular}{p{1.95cm}<{\centering}p{0.65cm}<{\centering}p{0.65cm}<{\centering}p{0.65cm}<{\centering}p{0.65cm}<{\centering}p{0.65cm}<{\centering}p{0.65cm}<{\centering} p{0.65cm}<{\centering}p{0.65cm}<{\centering}}
        \specialrule{0.1em}{0.6pt}{0pt}
        ${\beta}$ &0 & 0.0001 & 0.0005 & 0.001 & 0.005 & 0.01 & 0.05 & 0.1 \\
        \hline
        Indian Pines &64.40 &85.78 &92.90&94.06&\textbf{95.02}&94.19&94.64&94.59 \\
        University of Pavia &69.97& 91.93 & 95.77 & \textbf{96.50} & 96.38 & 94.40 & 93.41 & 93.63\\
        \toprule [1 pt]
    \end{tabular}
    \vspace{-0.25cm}
\end{table}

Besides, we further investigate the effect of the tuning parameter ${\beta}$ on the classification performance of HSIs. Table I reports the OA results for different ${\beta}$, from which we can observe that the proposed model can obtain the quasi-optimal performance when $\beta$ is set to 0.005 and 0.001 for these two HSI data sets, respectively.

\subsection*{B. Classification Performance}
The classification results of all methods on these two HSI data sets are reported in Tables II-III. We can find that the proposed method outperforms other methods. On the one hand, Sa-GCN and Se-GCN effectively learn the spatial and spectral correlations of HSI data. On the other hand, GCAFM adaptively explores complementary information of the extracted spatial and spectral features and suppresses noise interference in Sa-GCN and Se-GCN. From Tables II-III, compared with the second best classification performance achieved by
DIGCN or MDGCN, the OA results of ACSS-GCN-A on two data sets are increased by $1.00\%$ and $1.28\%$, respectively. In particular, the comparison between two fusion strategies reveals that ACSS-GCN-A can achieve the best classification results and bring at least $0.45\%$ increments in OA over the ACSS-GCN-C. In addition, the total sum of the training and testing computational time (seconds (s)) of all methods on two data sets is summarized in the last row of Tables II-III, from which compared with other GCN-based methods, the GCN method \cite{TN2017} consumes more time owing to the participation of all pixels in training step;
however, our method only costs a little more time to construct the spectral graph and run the GCAFM, but achieves the best performance for HSI classification.
\begin{table}[H]
   \centering
   \setlength{\abovecaptionskip}{-2pt}
   \renewcommand\tabcolsep{3.5pt}
   \label{tab:default}
   \caption{Classification Results for the Indian Pines Data Set}
    \scriptsize
   \begin{tabular}{p{0.65cm}<{\centering}p{0.60cm}<{\centering}p{0.60cm}<{\centering}p{0.60cm}<{\centering}p{0.60cm}<{\centering}p{0.80cm}<{\centering}p{0.60cm}<{\centering}p{0.60cm}<{\centering}p{0.80cm}<{\centering}p{0.80cm}<{\centering}}
        \specialrule{0.1em}{0pt}{0pt}
         \multirow{2}{*}{Class} & \multirow{2}{*}{SVM}  & 2-D   & \multirow{2}{*}{GCN}  & \multirow{2}{*}{FuNet}  &
         AMD-&
         MD-&
         DI-&
         {ACSS-} &
         ACSS- \\
         &&CNN&&&PCN&GCN&GCN&GCN-C&GCN-A\\
        \hline
        1 &\textbf{100.00}  &\textbf{100.00}
        &97.82
        &\textbf{100.00}
        & \textbf{100.00}
        & \textbf{100.00}
        &93.75
        &\textbf{100.00}
        & \textbf{100.00}\\
        2 &55.22       &58.58            &48.67     &60.52 &52.79            &78.68    &89.63      &\textbf{92.03}  & 89.49\\
        3 &80.50          &85.00            &64.10     &66.75   &57.87          & 94.88  &94.88       &95.86  &\textbf{96.38}\\
        4 &99.52     &\textbf{100.00}            &91.98     &90.34  &95.65           & 96.14    &96.16
        &\textbf{100.00}  & 99.61\\
        5 &77.04      &85.43            &89.86     &91.83   &88.18          & 90.73    &\textbf{98.45} &92.49  & 96.29\\
        6 &76.86   &71.57            &95.75     &\textbf{98.29 }  &96.78          & 97.86          &97.43 &97.57  & 96.11\\
        7 &\textbf{100.00} &\textbf{100.00}  &92.85 &\textbf{100.00}     &\textbf{100.00}            &\textbf{100.00}
        &69.23    &\textbf{100.00}  & 90.76\\
        8 &95.76        &96.42            &89.96     &99.78   &97.66          & \textbf{100.00}     &\textbf{100.00}       &\textbf{100.00}  & 99.77\\
        9 &\textbf{100.00}  &\textbf{100.00}
        &\textbf{100.00}    &\textbf{100.00}    &\textbf{100.00}            & \textbf{100.00} & \textbf{100.00} &\textbf{100.00}  &  \textbf{100.00}\\
        10 &56.47     &59.76            &69.75     &71.23  &72.13           & \textbf{94.16}    &89.92       &83.97 & 90.02\\
        11 &59.09     &64.12            &61.22     &56.53 &72.64            & \textbf{96.58}   &92.21        &94.22  &94.49\\
        12 &74.78      &76.38            &53.29     &70.52  &54.71           & 90.76   &84.55        &90.94  & \textbf{91.79}\\
        13 &98.86     &98.86            &97.56     &\textbf{100.00}
        & 99.42 &97.71 &97.14 &97.14  &\textbf{100.00}\\
        14 &87.61     &86.48            &85.30     &84.21 &86.07  & 99.68          &99.76 &\textbf{99.83}  & 99.66\\
        15 &94.94  &93.54            &46.63     &77.81 &75.84& \textbf{99.44}         &98.03 &98.60 & 98.82\\
        16 &\textbf{100.00} &\textbf{100.00}            &96.77     &\textbf{100.00} &\textbf{100.00} & \textbf{100.00}    &98.41      &\textbf{100.00}  & 99.68\\
        \hline
        OA &71.57      &73.94            &69.71     &72.93 &74.15  & 93.84 &93.76 &94.39  & \textbf{94.84}\\
        AA &84.79       &86.00            &80.10     &85.49 &84.38  & 96.04 &93.72 &96.41  & \textbf{96.43}\\
        \(\kappa\) &67.97  &70.50            &65.86     &69.45  &70.56 & 92.94 &92.86 &93.58 & \textbf{94.10}\\
         \hline
         Time(s) & 398.56 &      463.55  & 137.23   &   53.67  &    57.17       &32.38   &   23.17   &     86.69   &   82.14\\
        \toprule [1 pt]

    \end{tabular}
    \vspace{-0.25cm}
\end{table}

\begin{table}[H]
    \centering
    \setlength{\abovecaptionskip}{-2pt}
    \renewcommand\thetable{\Roman{table}}
    \renewcommand\tabcolsep{3.0pt}
    \caption{Classification Results for the University of Pavia Data Set}
     \scriptsize
       \begin{tabular}{p{0.65cm}<{\centering}p{0.65cm}<{\centering}p{0.70cm}<{\centering}p{0.70cm}<{\centering}p{0.65cm}<{\centering}p{0.75cm}<{\centering}p{0.65cm}<{\centering}p{0.65cm}<{\centering}p{0.80cm}<{\centering}p{0.80cm}<{\centering}}
        \specialrule{0.1em}{0pt}{0pt}
         \multirow{2}{*}{Class} & \multirow{2}{*}{SVM}  & 2-D   & \multirow{2}{*}{GCN}  & \multirow{2}{*}{FuNet}  &
         AMD-&
         MD-&
         DI-
         & {ACSS-} & ACSS- \\
         &&CNN&&&PCN&GCN&GCN&GCN-C&GCN-A\\
        \hline
         1          &63.29           &58.86       &74.00        &82.85  & 81.30& \textbf{92.27}  &88.50        &88.35  & 86.12\\
        2           &62.91      &77.79       &61.95        &93.87 &97.89          &94.44     &91.78     &97.55  & \textbf{98.96}\\
        3           &41.32        &52.58       &62.12        &79.07  &76.94         & 88.88         &94.20 &\textbf{99.85}  &99.08\\
        4           &71.65       &79.23       &95.52        &\textbf{95.81} &91.33       & 94.26    &91.76 &92.04  & 92.35\\
        5           &83.65          &88.06       &97.62         &99.70  &\textbf{99.77}         & 99.01    &99.39 &99.01  & 99.01\\
        6           &51.09       &54.51       &44.87         &71.51   &67.87       & \textbf{100.00}   &\textbf{100.00}       &\textbf{100.00}  & \textbf{100.00}\\
        7           &51.53     &61.76       &82.85         &86.77    &84.61      &\textbf{99.69}    &\textbf{99.69} &99.63  & 97.83\\
        8           &55.78       &76.59       &85.25         &70.45   &71.93       & 98.49           &\textbf{99.45} &96.82  & 98.14\\
        9           &88.66   &94.33  & 98.83 &99.34 & \textbf{99.89} &96.51 &87.68 &88.85  &  96.07\\
        \hline
        OA         &61.39          &71.07       &68.81         &87.02 &87.76  & 95.17 &93.40 &95.99  & \textbf{96.45}\\
        AA         &63.32        &71.52      &78.11         &86.60 &85.73 & 95.95 &94.72 &95.79  & \textbf{96.40}\\
        \(\kappa\) &51.13      &61.22       &58.38         &82.76 &60.77 & 93.65 &91.38 &94.69 & \textbf{95.30}\\
         \hline
         Time(s) & 163.16 &     1849.85 &  2093.78    &61.63 &      62.00 &    164.31  &    49.70  &     271.26  &    266.06\\
        \toprule[1pt]
    \end{tabular}
    \vspace{-0.25cm}
\end{table}

\begin{figure}[ht!]
	\tiny
	\centering
    \mbox{
        \subfigure[\label{subfig:isp}]{\includegraphics[width=0.15\linewidth]{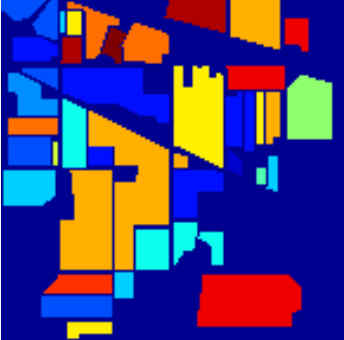}}
        \subfigure[\label{subfig:isp}]{\includegraphics[width=0.15\linewidth]{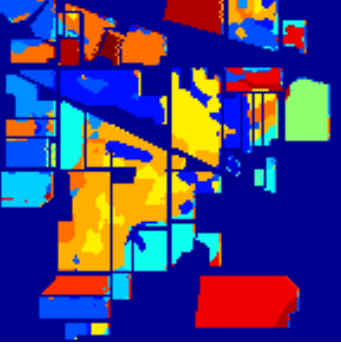}}
        \subfigure[\label{subfig:isp}]{\includegraphics[width=0.15\linewidth]{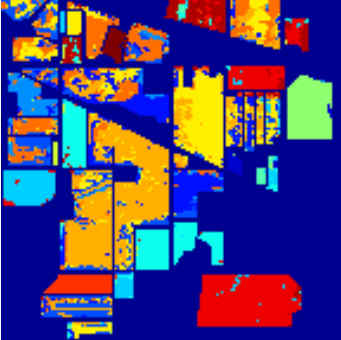}}
        \subfigure[\label{subfig:isp}]{\includegraphics[width=0.15\linewidth]{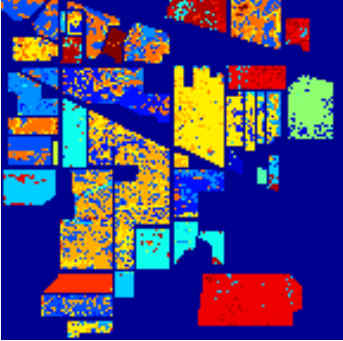}}
        \subfigure[\label{subfig:isp}]{\includegraphics[width=0.15\linewidth]{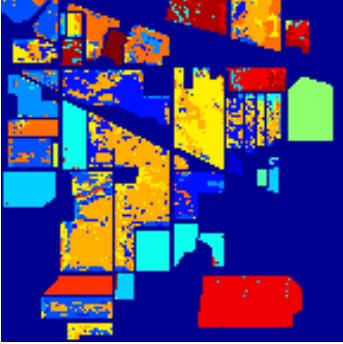}}
	
}\\
	\mbox{
		\subfigure[\label{subfig:isp}]{\includegraphics[width=0.16\linewidth]{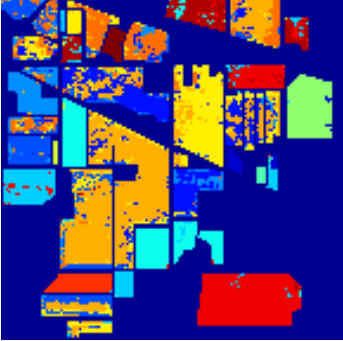}}
        \subfigure[\label{subfig:isp}]{\includegraphics[width=0.15\linewidth]{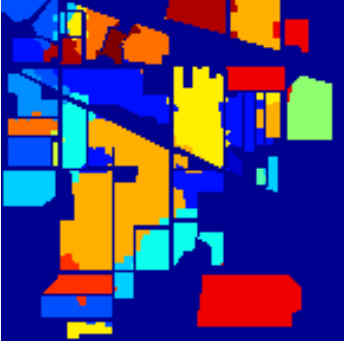}}
        \subfigure[\label{subfig:isp}]{\includegraphics[width=0.15\linewidth]{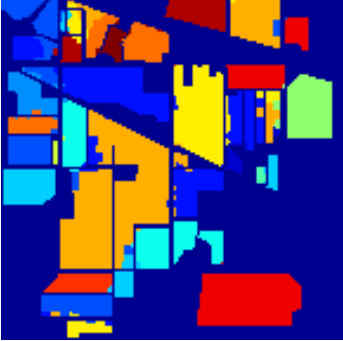}}
        \subfigure[\label{subfig:isp}]{\includegraphics[width=0.15\linewidth]{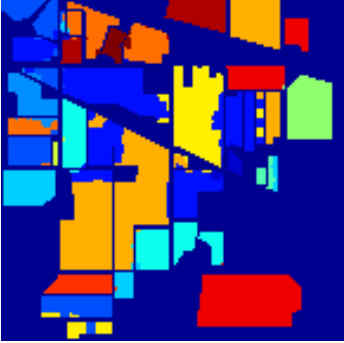}}
        \subfigure[\label{subfig:isp}]{\includegraphics[width=0.15\linewidth]{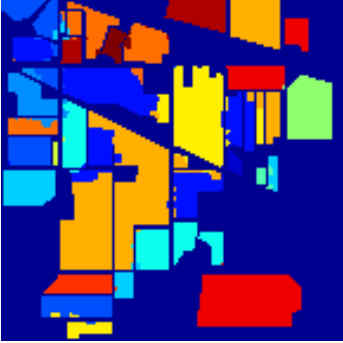}}
	}
	\caption{Classification maps for the Indian Pines Data Set. (a) Ground-truth map.  (b) SVM.  (c) 2-D CNN. (d) GCN.  (e) FuNet. (f) AMDPCN. (g) MDGCN. (h) DIGCN. (i) ACSS-GCN-C.  (j) ACSS-GCN-A.}
	\label{fig:hh_are}
\end{figure}

To analyze the classification performance more intuitively, the visualized results of all methods on the Indian Pines data set are shown in Fig. 2, from which it can be observed that the classification map of our proposed model is closest to the ground-truth map,
and there are least noise and misclassifications. Particularly, for the GCN, FuNet, and ADMPCN methods, the pepper-noise-like mistakes in certain regions can be observed due to the loss of spatial context information, leading to poor classification results. The above experimental results further illustrate the superiority of ACSS-GCN.

\subsection*{C. Sensitivity Comparison and Analysis under Small Samples}
To further illustrate the classification performance under the small number of training samples, 5, 10, 15, 20, 25, and 30 samples of each class are randomly selected from these two HSI data sets for training. Fig. 4 shows the OA curves with different numbers of training samples. As expected, the performance of all methods is improved with the increase of training samples. The ACSS-GCN method outperforms the other methods, which verifies the advantages of our ACSS-GCN method. It is noted that the ACSS-GCN-A is superior to the ACSS-GCN-C model in most cases, which may be because the concatenation operation generate more redundant features than the addition operation affecting the final classification performance. These experimental results also show that the ACSS-GCN model is superior to other considered models.

\begin{figure}[h]
    \centering
    \renewcommand\tabcolsep{1.0pt}
    \scriptsize
    \begin{tabular}{cc}
        \begin{minipage}[t]{0.35\linewidth}
            \centering
            \includegraphics[height=1.1in,width=1.4in]{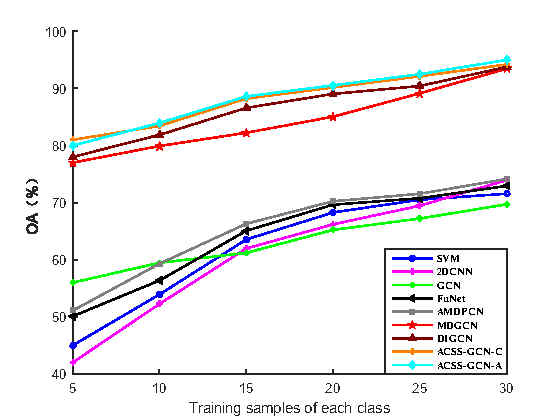}
         \end{minipage}
         &\begin{minipage}[t]{0.5\linewidth}
            \centering
            \includegraphics[height=1.1in,width=1.4in]{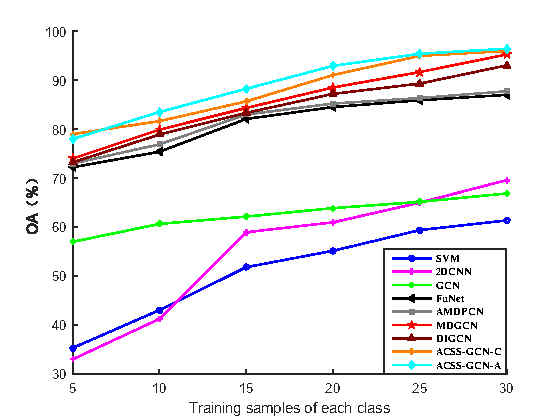}
        \end{minipage}
        \\
        (a) Indian Pines& (b) University of Pavia  \\
    \end{tabular}
\caption{OA (\%) of all methods with different number of training samples for two HSI data sets. (a) Indian Pines. (b) University of Pavia.}
\end{figure}

\subsection*{D. Ablation Study}
To highlight the effectiveness of Se-GCN, Sa-GCN, and GCAFM in our ACSS-GCN-A, detailed ablation studies are conducted to see how they contribute to the classification accuracy, where the ASS-GCN-A is a variant of ACSS-GCN-A by replacing the GCAFM with the element-wise addition operation. From Table IV, by directly adding the features of Se-GCN and Sa-GCN, the ASS-GCN-A can extract the joint spatial-spectral features for good classification performance of HSIs. Furthermore, by introducing the idea of cross attention, a GCAFM is designed to fully explore the complementary information between Se-GCN and Sa-GCN. Compared with the ASS-GCN-A, GCAFM can bring 0.76\% and 0.49\% gains to the ACSS-GCN-A for these two data sets, respectively. Experimental results show that the Se-GCN and Sa-GCN are effective and complementary, and the joint learning of spatial-spectral information can further improve the performance of HSI classification. Besides, the GCAFM can be benefit for improving the classification accuracy of the ACSS-GCN.
  \begin{table}[H]
    \centering
    \setlength{\abovecaptionskip}{-2pt}
    \renewcommand\thetable{\Roman{table}}
    \renewcommand\tabcolsep{3.0pt}
    \caption{Ablation Studies of ACSS-GCN on the Indian Pines and University of Pavia Data Sets}
    \scriptsize
    \begin{tabular}{p{0.80cm}<{\centering} |p{0.65cm}<{\centering} p{0.65cm}<{\centering} p{0.80cm}<{\centering} p{0.80cm}<{\centering}|p{0.65cm}<{\centering} p{0.65cm}<{\centering} p{0.80cm}<{\centering} p{0.80cm}<{\centering}}
        \specialrule{0.1em}{0pt}{0pt}
         \hline
         \multirow{3}{*}{Methods} & \multicolumn{4}{c|}{Indian Pines}& \multicolumn{4}{c}{University of Pavia} \\
        &Se-& Sa- &  ASS-&ACSS-
        & Se-& Sa- &  ASS-&ACSS- \\
         &GCN&GCN&GCN-A&GCN-A
         &GCN&GCN&GCN-A&GCN-A\\

        \hline
        OA&50.52&93.87&94.08&\textbf{94.84} &43.54&95.69&95.96&\textbf{96.45}\\
        \(\kappa\)&44.82&92.98&93.23&\textbf{94.10}
        &33.52&94.29&94.64&\textbf{95.30}\\
        \toprule[1pt]
    \end{tabular}
\end{table}

\section{Conclusion}
In this letter, a novel ACSS-GCN framework has been proposed for HSI classification. Firstly, a dual-branch GCN-based spatial-spectral structure is proposed as a backbone of ACSS-GCN to jointly learn the spatial and spectral information of HSI data. Then, the GCAFM is designed to explore complementary
information of Sa-GCN and Se-GCN for better HSI classification. Finally, an adaptive graph is proposed to dynamically update the spatial and spectral
graphs during the back propagation of the whole model. Experimental results on two HSI data sets show that our method offers better classification performance than other GCN-based methods.

\ifCLASSOPTIONcaptionsoff
  \newpage
\fi

\end{document}